\documentclass{article}

\usepackage[final]{neurips_2020}

\usepackage[utf8]{inputenc} 
\usepackage[T1]{fontenc}    
\usepackage{url}            
\usepackage{booktabs}       
\usepackage{amsfonts}       
\usepackage{nicefrac}       
\usepackage{microtype}      
\usepackage{xcolor}         

\usepackage{enumitem}
\usepackage{graphicx}
\usepackage{multirow}
\usepackage{amsmath,amssymb} 
\usepackage[colorlinks,citecolor=blue,urlcolor=magenta,bookmarks=false,hypertexnames=true]{hyperref}

\title{Crowd Density Estimation using Imperfect Labels}

\author{%
  Muhammad Asif Khan\thanks{Corresponding author} \\
  Qatar Mobility Innovations Center (QMIC)\\
  Qatar University\\
  Doha, Qatar \\
  \texttt{mkhan@qu.edu.qa} \\
  \And
  Hamid Menouar \\
  Qatar Mobility Innovations Center (QMIC) \\
  Qatar University\\
  Doha, Qatar \\
  \texttt{hamidm@qmic.com} \\
  \And
  Ridha Hamila \\
  Department of Electrical Engineering \\
  Qatar University\\
  Doha, Qatar \\
  \texttt{hamila@qu.edu.qa} \\
}

\begin{document}
\maketitle

\begin{abstract}
Density estimation is one of the most widely used methods for crowd counting in which a deep learning model learns from head-annotated crowd images to estimate crowd density in unseen images. Typically, the learning performance of the model is highly impacted by the accuracy of the annotations and inaccurate annotations may lead to localization and counting errors during prediction. A significant amount of works exist on crowd counting using perfectly labeled datasets but none of these explore the impact of annotation errors on the model accuracy. In this paper, we investigate the impact of imperfect labels (both noisy and missing labels) on crowd counting accuracy. We propose a system that automatically generates imperfect labels using a deep learning model (called annotator) which are then used to train a new crowd counting model (target model). Our analysis on two crowd counting models and two benchmark datasets shows that the proposed scheme achieves accuracy closer to that of the model trained with perfect labels showing the robustness of crowd models to annotation errors.
\end{abstract}

\section{Introduction} \label{sec:intro}
Crowd counting and density estimation is an important problem with many interesting applications. For instance, crowd counts in public places such as political rallies, stadiums, and exhibition centers can be of great interest to authorities and event organizers for effective management and control actions. Vehicle counting is significant in transport planning, road intersection design, traffic signal control, and emergency vehicle preemption strategies. Similarly, wildlife counting in forests can play a role to protect species of interest.

The diverse applications of crowd density estimation and counting has attracted the computer vision community. Over the past years, several state-of-the-art deep learning models using convolution neural networks (CNNs) have been proposed \cite{Khan2022RevisitingCC, TAFNET_2022, SGANet_IEEEITS2022, SASNet_AAAI2021, DeepCount_ECAI2020, TEDnet_CVPR2019, CANNet_CVPR2019, CSRNet_CVPR2018, GSP_CVPR2019, SANet_ECCV2018}. These models are trained with accurately labeled data (i.e., dot annotated images). However, the acquisition of dot-annotated data can be expensive for dense crowds and a large number of images. Moreover, the annotations may not be available beforehand in some scenarios due to security and privacy reasons.

In this paper, we propose an intriguing approach in which the network first generates density maps (imperfect and noisy) labels and then trains a target model on the generated labeled data. The motivation for the proposed approach is two-fold. First, dot annotations are expensive and thus the automatic generation of ground truth density maps saves time. Second, the perfect labels may not be available in some scenarios beforehand. For instance, deploying several drones (each running a lightweight model) for the prediction can simultaneously fine-tune the model weights over time using federated learning \cite{Liu_2022, Osifeko_2021, unal2021integration}. However, the drones will need labels to predict the count in the crowd images. Thus, one can use the proposed method to use a secondary deep model to predict the density map for the image which will serve as a ground truth to train the primary (target) model. The deep model will only be used for inference and when the node is participating in the federating learning round. Over time, the lightweight target model will be fine-tuned and its accuracy will improve.

The contribution of the paper is as follows: We propose a novel method for training a lightweight crowd counting model by automatic label generation using a secondary deep crowd counting model (annotator). The auto-generated labels are typically imperfect and noisy. Whether such noisy and imperfect labels can be used to train another model? We trained two different models of different sizes and architectures on two benchmark datasets to evaluate the efficacy of the proposed method. The performance is investigated using standard metrics and some intriguing results are achieved.

\section{Related Work} \label{sec:rel_work}
Density estimation using CNN for crowd counting was first proposed in \cite{CrowdCNN_CVPR2015}. The authors propose a simple single-column architecture with six layers. Several other works followed the approach and proposed different CNN architectures of various sizes typically to improve accuracy over benchmark datasets. The architectures  used in these works include multi-column CNN \cite{MCNN_CVPR2016, CrowdNet_CVPR2016, SCNN_CVPR2017, CMTL_AVSS2017}, modular CNNs \cite{MSCNN_ICIP2017, SANet_ECCV2018, SGANet_IEEEITS2022}, encoder-decoder networks \cite{TEDnet_CVPR2019, MobileCount_PRCV2019}, and transfer-learning based models \cite{CSRNet_CVPR2018, CANNet_CVPR2019, GSP_CVPR2019, TAFNET_2022}.
The initial small-sized single-column architectures (e.g. \cite{CrowdCNN_CVPR2015} generally achieve poor accuracy on images having large scale variations. Scale variations typically arise from the perspective distortions and different resolution images in the dataset. Thus, multi-column networks with filters of varying receptive fields in different columns can be used to capture the scale variations. A multi-column convolution neural network (MCNN) is proposed in \cite{MCNN_CVPR2016}. MCNN is a three-column architecture with variable sizes filters ($9\times9, 7\times7, 5\times5, \text{and } 3\times3$) in each column. The switching-CNN \cite{SCNN_CVPR2017} addresses the scale variations differently. It uses three CNN networks (regressors) to process the input image using only one column based on the density of the image. The density is automatically estimated by another single-column CNN (classifier or switch). One shortcoming of multi-column networks is their capability to adapt to the scale variations which is limited by the number of columns i.e., more columns are needed when there are large variations in the images across the dataset. An alternative solution is to use modular networks which typically consist of a single-column (sometimes multiple columns) architecture with scale-adaptive feature extraction modules. These models are mainly inspired by the Inception-like models \cite{Inception_CVPR2015}. Encoder-decoder models \cite{TEDnet_CVPR2019, MobileCount_PRCV2019} are also being used in many research works. These models follow the popular UNet architecture \cite{UNet_2015}, where the encoder part first learns and extracts features from the input image, and then the decoder part generates prediction using the features passed by the encoder. Encoder-decoder models produce high-quality density maps. Crowd counting in dense images and congested scenes can become more challenging. Thus, a significant amount of research contributions adopt a transfer learning approach i.e., to use a pre-trained image classification model e.g., VGG-16 \cite{VGG16_ICLR2015}, ResNet \cite{ResNet_CVPR2016} or Inception \cite{Inception_CVPR2015} as a front-end (or backbone) for feature extraction and a shallow CNN network (back-end) to use the features for estimating the crowd density. These models are generally more accurate and faster to train but incur longer inference delays at prediction.
\par
The aforementioned crowd counting models use accurately dot annotated localization maps as ground truth for the crowd images to train the model. All these works focus on investigating the accuracy improvement, but none of these explore the model performance over imperfectly labeled data.

\section{Proposed Scheme} \label{sec:scheme}
\begin{figure*}[htp]
    \centering
    \includegraphics[width=0.85\textwidth]{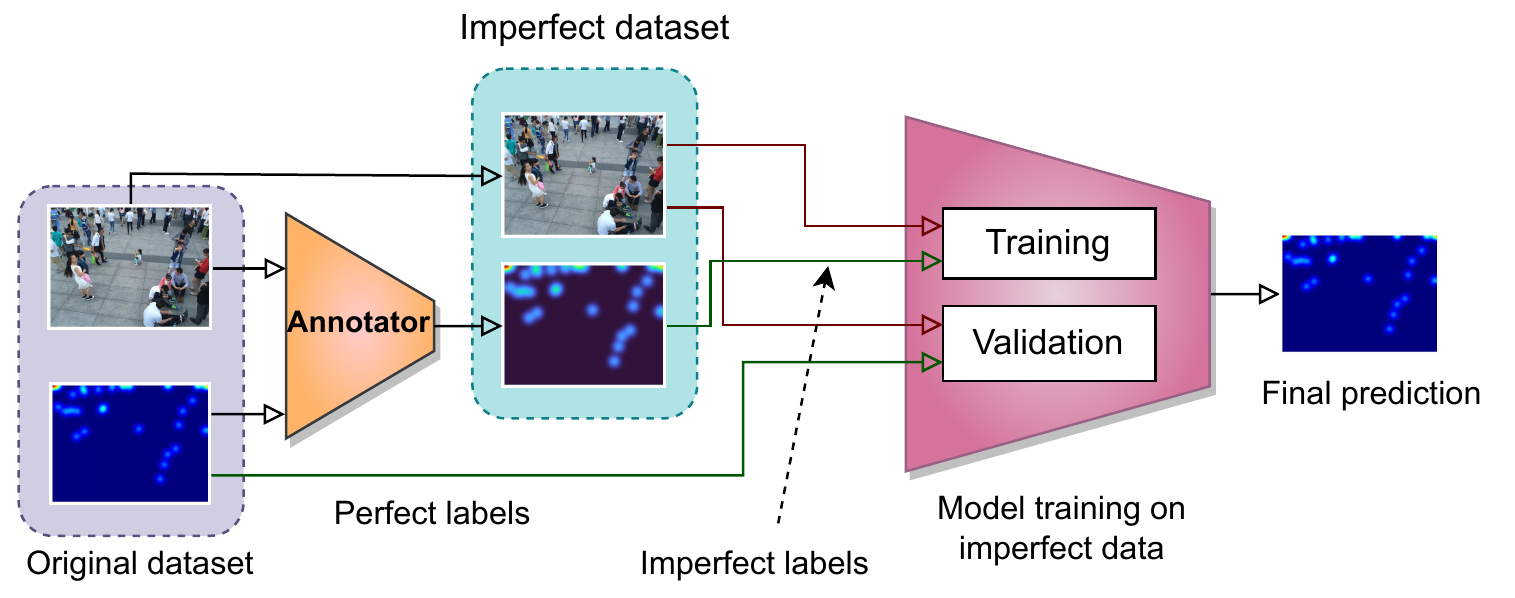}
    \caption{Proposed method for automatic generation of ground truth and training a lightweight crowd counting model using imperfect (noisy annotation) dataset.} \label{fig:prop_scheme}
\end{figure*}

The proposed scheme is illustrated in Fig. \ref{fig:prop_scheme}. In the first step, the system generates imperfect labels for the entire dataset using a deep network model (Annotator). The predictions using the annotator network serve as imperfect labels for training the target network. In the next step, the target network is trained on the noisy (imperfect labels). During the training process, the model learns from the imperfect labels by computing the loss function which is $l_2$ distance between the imperfect labels and the prediction of the target network. Then, to find the accuracy of the model, the mean absolute error (MAE) or $l_1$ distance between the predictions of the target network and the original ground truth (perfect) is calculated. In this way, the model learns from the imperfect data but its accuracy is tested on the actual ground truth.

\begin{figure} \centering
\includegraphics[width=\columnwidth]{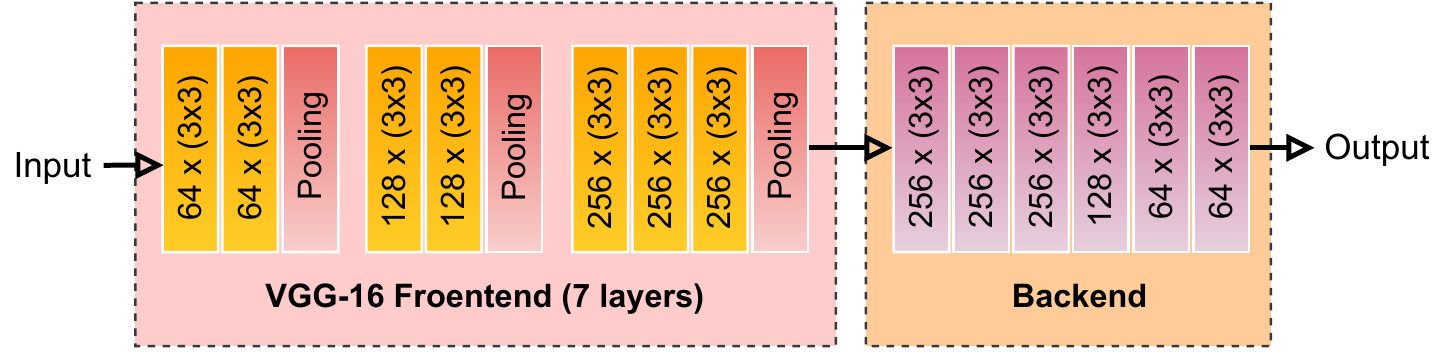}
\caption{CSRNet\_lite: A modified lightweight version of CSRNet \cite{CSRNet_CVPR2018}.}
\label{fig:CSRNet_lite}
\end{figure}

We choose the CSRNet \cite{CSRNet_CVPR2018} model as the annotator network due to its good accuracy. However, CSRNet generates density maps of size 1/8 of the input image, whereas the target network used in our study generates density maps of size 1/4 of the input image. To solve this issue, we modified the original CSRNet \cite{CSRNet_CVPR2018} architecture. CSRNet uses the first 10 layers of VGG-16 \cite{VGG16_ICLR2015} network. In VGG-16, pooling layers are used to reduce the size of the predicted density map to half after layer-2 (1/2), layer-4 (1/4), layer-7 (1/8), and layer-10 (1/16). Using the first 10 layers (without pooling layer after layer-10), CSRNet generates an output of size 1/8. In our modified version, We took only the first seven (7) layers (without the pooling layer after layer-7) to generate an output of size 1/4. Furthermore, the back-end network of CSRNet uses layers of sizes (512, 512, 512, 256, 128, 64), respectively. We also modified the back-end network to reduce layers sizes to (256, 256, 256, 128, 64, 64) to match the number of channels in the output of the front-end and input of the back-end networks. We denote the modified CSRNet architecture as CSRNet\_lite (due to its relatively smaller size). It is shown in Fig. \ref{fig:CSRNet_lite}.

The CSRNet\_lite has $3.9$ Million parameters, almost four times less than CSRNet ($16.2$ Million).

\subsection{Model Training}
The proposed scheme is evaluated over the DroneRGBT dataset \cite{DroneRGBT_dataset}. The DroneRGBT dataset has 1807 RGB and thermal image pairs in the train set. Each image has a spatial resolution of 512×640 pixels. We split the train set into a ratio of (70\% : 30\%) for training and validation. The dataset covers several scenes (e.g., campus, streets, public parks, car parking, stadiums, and plazas) and contains diverse crowd densities, illumination, and scale variations. The dataset provides head annotations of people.

To train CSRNet\_lite model to generate imperfect labels, the dot annotations are first converted to ground truth (GT) density maps (perfect labels). To generate the GT density map the head positions ($x_i$) expressed as delta functions $\delta(x - x_i)$ are convolved with Gaussian kernels ($G_\sigma$).
\begin{equation}
    D_i^{gt} = \sum_{i=1}^{N}{ \delta(x-x_i) * G_\sigma}
\end{equation}
where, N denotes the total number of dots (i.e., head positions) in the image. The value of $\sigma=7$ is empirically determined by analyzing different values and visually seeing the overlapping between the blobs in the GT density maps. To prevent overfitting, we applied three different data augmentations techniques including random brightness, random contrast, and horizontal flipping. We use Adam optimizer \cite{Adam_ICLR2015} with an initial learning rate $0.0001$ to optimize the loss function. To compute the loss, we use the pixel-wise Euclidean distance between the target and predicted density maps as in Eq. \ref{eq:mse}.

\begin{equation} \label{eq:mse}
    L(\Theta) = \frac{1}{N} \sum_{1}^{N}{ ||D(X_i;\Theta) - D_i^{gt}||_2^2}.
\end{equation}

where $N$ is the number of heads in the image, $D(X_i;\Theta)$ is the model prediction density map and $D_i^{gt}$ is the GT density map. The predicted density maps are then used to generate another dataset with all the images of the original dataset whereas ground truth labels are replaced by the predictions from CSRNet\_lite. This noisy dataset is then used to train a lightweight model. We use two lightweight models i.e., MCNN \cite{MCNN_CVPR2016}, and LCDnet \cite{LCDnet_2022} to train from scratch using the noisy datasets. The same settings of optimizer and augmentation are used in the second stage as well. We use a single machine with two Nvidia RTX-8000 GPUs and PyTorch deep learning framework for training the model.

\section{Evaluation and Results} \label{sec:results}

\subsection{Evaluation Metrics} \label{subsec:metrics}
We evaluate the proposed method using the most widely used metrics i.e., mean absolute error (MAE), Grid Average Mean Error (GAME), Structural Similarity Index (SSIM), and Peak Signal-to-Noise Ratio (PSNR) calculated in Eq. \ref{eq:mae}, \ref{eq:game}, \ref{eq:ssim}, and \ref{eq:psnr}, respectively. MAE and GAME measure the accuracy of the model by measuring the counting errors in GT and predicted density maps. The GAME metric is more sensitive to localization errors because it calculates MAE over patches of the predictions. The SSIM and PSNR measure the quality of the predicted density maps.

\begin{equation} \label{eq:mae}
    MAE = \frac{1}{N} \sum_{1}^{N}{(e_n - \hat{g_n})}.
\end{equation}
where, $N$ is the total number of images in the dataset, $g_n$ is the ground truth (actual count) and $\hat{e_n}$ is the prediction (estimated count) in the $n^{th}$ image.

\begin{equation} \label{eq:game}
    GAME = \frac{1}{N} \sum_{n=1}^{N}{ ( \sum_{l=1}^{4^L}{|e_n^l - g_n^l|)}}.
\end{equation}
We set the value of $L=4$, thus each density map is divided into a grid size of $4\times4$ creating $16$ patches.

\begin{equation} \label{eq:ssim}
    SSIM (x,y) = \frac{(2\mu_x \mu_y + C_1)  (2\sigma_x \sigma_y C_2)}  {(\mu_z^2 \mu_y^2 + C_1)  (\mu_z^2 \mu_y^2 + C_2)}.
\end{equation}
where $\mu_x, \mu_y, \sigma_x, \sigma_y$ represent the means and standard deviations of the actual and predicted density maps, respectively.

\begin{equation} \label{eq:psnr}
    PSNR = 10 log_{10}\left( \frac{Max(I^2)}{MSE}  \right).
\end{equation}
where $Max(I^2)$ is the maximal in the image data (I). If it is an 8-bit unsigned integer data type, the $Max(I^2)=255$.

\subsection{Results} \label{subsec:results}
\begin{figure}[!h] 
\centering
\includegraphics[width=0.99\columnwidth]{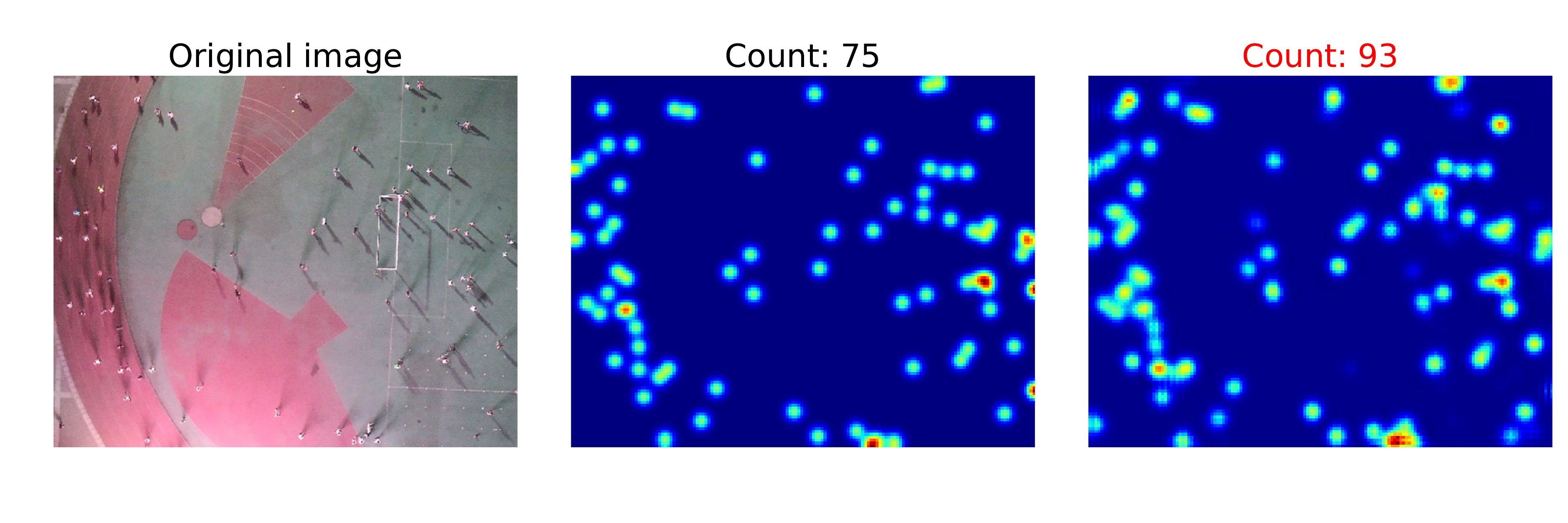}
\caption{The figure shows an image (left), its ground truth density map or perfect label (middle) generated from the dot annotation, and a predicted density map as the imperfect label (right) generated using the CSRNet\_lite model. The imperfect labels are used to train the lightweight model.}
\label{fig:imperfect_labels}
\end{figure}

The first step in the proposed method is the generation of imperfect labels. Ideally, such imperfect labels should not have large errors. A sample label generated using the CSRNet\_lite model for the given image and the ground truth density map is shown in Fig. \ref{fig:imperfect_labels}.

\begin{figure*}[htbp]
\centering
\includegraphics[width=0.99\textwidth]{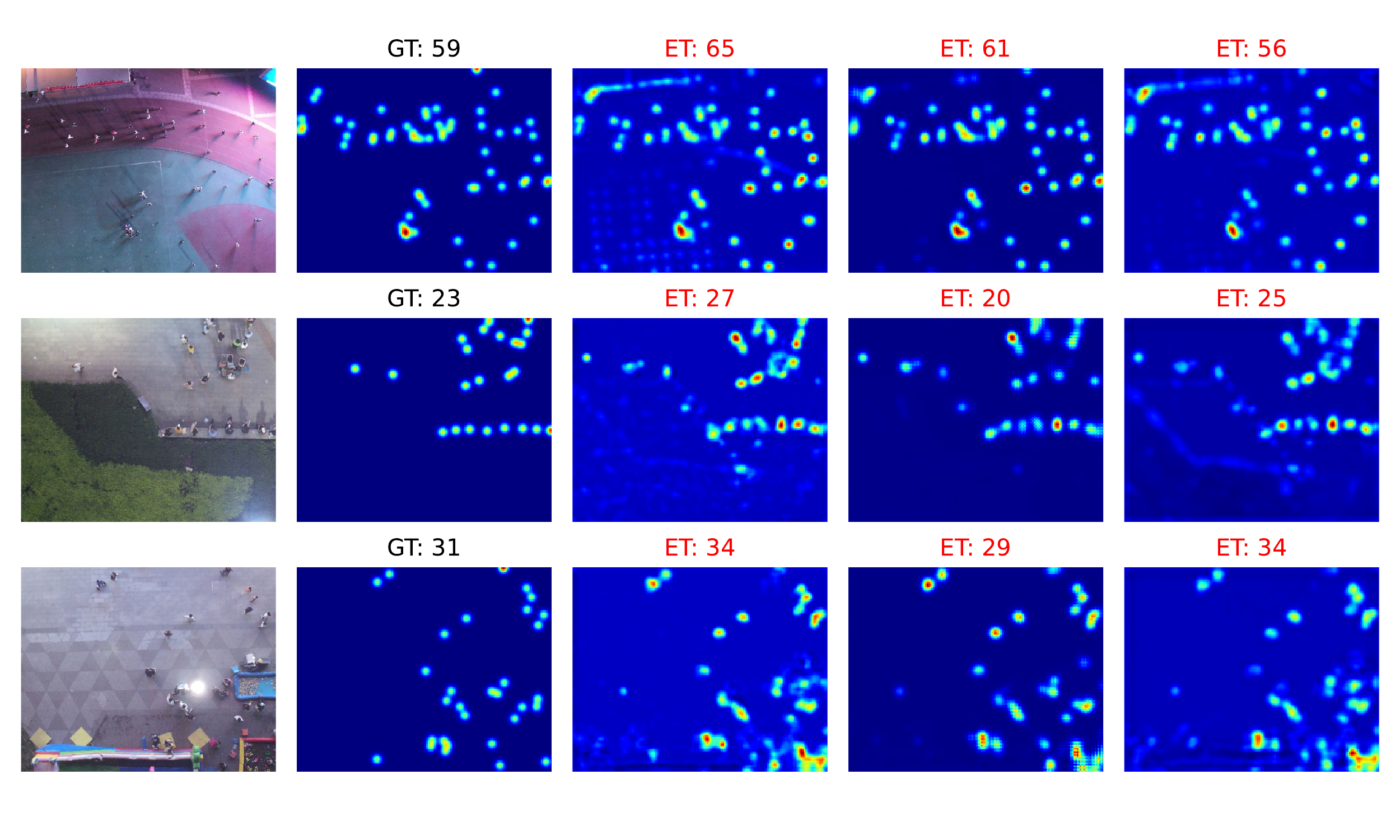}
\caption{Sample predictions on DroneRGBT dataset \cite{DroneRGBT_dataset}: Column 1 shows same images from the dataset. Column 2 shows the ground truth (perfect). Column 3 shows predictions of the model trained on the perfect ground truth. Column 4 shows imperfect labels (generated using CSRNet\_2 network. Column 5 shows predictions using the model trained on imperfect data. GT means ground truth (actual count), whereas ET denotes estimated truth (predicted count).}
\label{fig:predictions}
\end{figure*}

It can be observed that the noisy label is not perfect but is reasonably accurate to aid in training the target model. The performance of the two target models (i.e., MCNN and LCDnet) trained using the original target data (perfect data) and on the noisy dataset is presented in Table \ref{tab:performance} to show the learning performance using imperfect data.

\begin{table*}[!h] \centering
\caption{Performance analysis of the proposed method using two lightweight models (MCNN and LCDnet) over two datasets (DroneRGBT, CARPK).}
\label{tab:performance}
\begin{tabular}{r|r|r|cccc} \toprule
Model   &Dataset  &Labels &MAE  &GAME  &SSIM  &PSNR \\[1em] \midrule \midrule

\multirow{2}{*}{MCNN}
& \multirow{2}{*}{DroneRGBT}
&Perfect &16.88 &43.49 &0.66 &23.47   \\[0.5em]
& &Imperfect &17.86 &46.92 &0.64 &23.07   \\[1.5em] \midrule

\multirow{2}{*}{MCNN}
& \multirow{2}{*}{CARPK}
&Perfect &10.10 &42.40 &0.76 &19.21   \\[0.5em]
& &Imperfect &10.86 &43.92 &0.64 &18.03   \\[1.5em] \midrule

\multirow{2}{*}{LCDnet}
& \multirow{2}{*}{DroneRGBT}
&Perfect &21.40 &46.92 &0.62 &21.39   \\[0.5em]
& &Imperfect &23.82 &47.91 &0.60 &20.07   \\[1.5em] \midrule

\multirow{2}{*}{LCDnet}
& \multirow{2}{*}{CARPK}
&Perfect &13.10 &46.13 &0.69 &20.14   \\[0.5em]
& &Imperfect &13.75 &48.26 &0.66 &20.07   \\[1em]

\bottomrule
\end{tabular}
\end{table*}

One can observe that both networks (MCNN and LCDnet) when trained over imperfect labels show very good accuracy on test (unseen) data. There is a very small or sometimes negligible difference in the MAE and GAME metrics for both models when trained on the two datasets using perfect and imperfect datasets, respectively. Fig. \ref{fig:predictions} show sample prediction over DroneRGBT \cite{DroneRGBT_dataset} dataset using the proposed scheme (showing both predictions over perfect and noisy datasets used in training). On most images, the predicted count was exactly the same for both predictions (using models trained over perfect and noisy data). This means, that even imperfect data with missing or noisy annotations (due to prediction errors) can be used by the model to learn features that are even not learned by the annotator network.

\begin{figure}[!h]
\centering
\includegraphics[width=0.8\columnwidth]{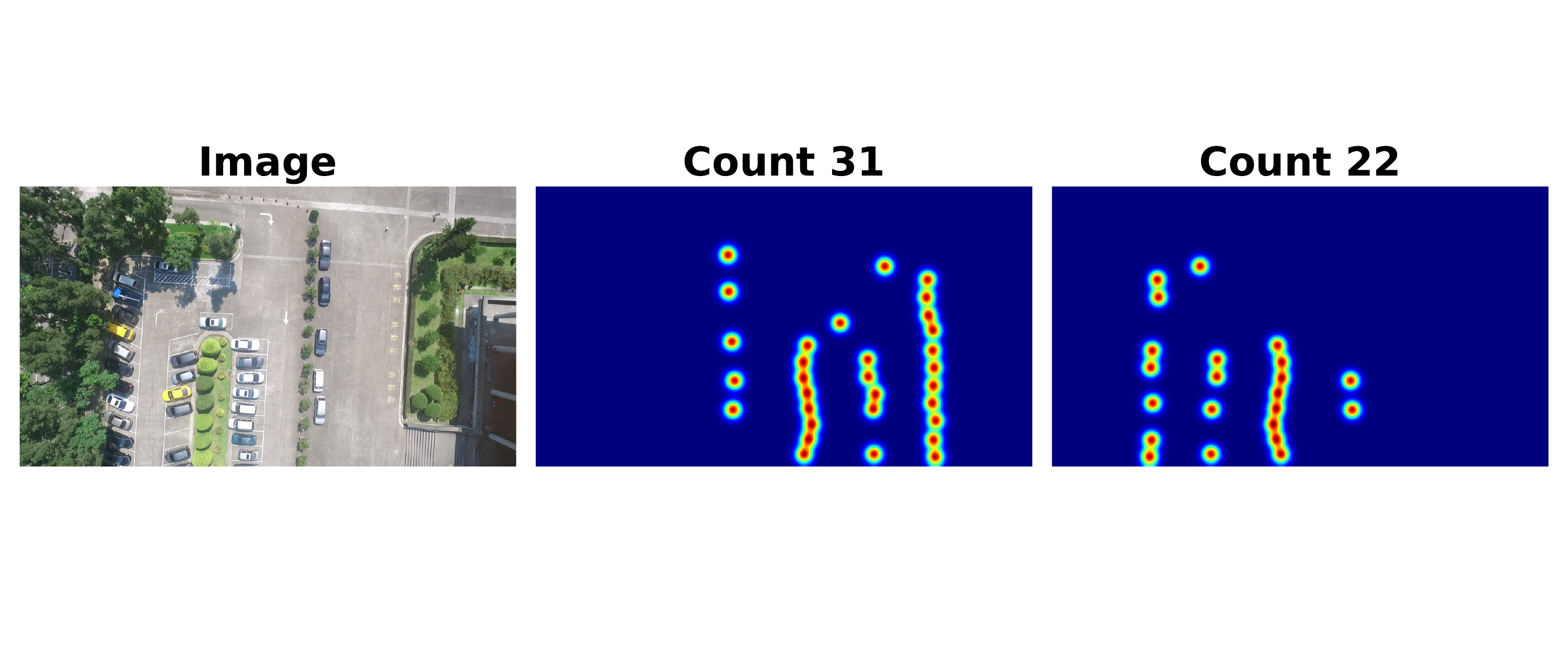}
\caption{A sample ground truth (GT) density map with missing (30\%) data.} \label{fig:missing_labels}
\end{figure}

\subsection{Ablation Study} \label{subsec:ablation}
Above we discussed the model learning performance over noisy or imperfect training data (generated by another model). The analysis shows that the model can learn features (objects) that are even not annotated in the training data from other annotations and similar features. To further investigate the learning process, we train and evaluate the two models using missing annotations instead of imperfect data (noisy labels). To implement this, we randomly delete a fixed portion (30\%) of annotations from each image to generate imperfect labels (see Fig. \ref{fig:missing_labels}).

\begin{table}[!h]
\centering
\caption{Ablation study using imperfect data (missing labels).} \label{tab:ablation}
\begin{tabular}{r|r|r|cc} \toprule
Model &Dataset &Annotation &MAE &GAME \\[1em] \midrule \midrule

\multirow{2}{*}{MCNN} & \multirow{2}{*}{CARPK} &Perfect &10.1 &43.4 \\[1em]
&&Missing labels &14.4 &53.6 \\[1.5em] \midrule

\multirow{2}{*}{LCDnet} & \multirow{2}{*}{CARPK} &Perfect &13.7 &48.2 \\[1em]
&&Missing labels &18.5 &55.3 \\[1em]
\bottomrule
\end{tabular}
\end{table}

The two models MCNN and LCDnet are then trained over this dataset with missing labels. The model performance is compared with the same model trained with perfect data. As anticipated, the model could learn unlabelled objects of interest. A comparison over CARPK \cite{CARPK_dataset} dataset is shown in Table \ref{tab:ablation}.

The results in Tab;e \ref{tab:ablation} strengthen the idea that having a good model and efficient training strategy can overcome the annotation errors (missing and noisy annotations) in the training data.

\section{Conclusion} \label{sec:conclusion}
The performance of any deep learning model is hugely correlated with the quality of the data. More specifically, accurate labels improve the learning process whereas noisy or imperfect labeled data can lead to inaccurate predictions. In crowd density estimation, the ground truth labels contain dot annotations of all heads (or object centers) in the image. Dot annotation is expensive and hence we propose to automatically generate imperfect labels by using a deep CNN model and then investigate the learning performance of a shallow model on the imperfect data. The results report that even imperfect data can be efficiently used to train another model from scratch. The proposed scheme can be efficiently used in fine-tuning crowd counting models running over distributed devices (such as drones) in a federated learning environment without human-aided labeling. In future work, we plan to apply the proposed method in other computer vision-based applications such as object detection and anomaly detection.
\par

\bibliographystyle{plainnat}
\bibliography{biblio}







\end{document}